\documentclass[sn-mathphys,Numbered]{sn-jnl}

\usepackage{graphicx}
\usepackage{multirow}
\usepackage{amsmath,amssymb,amsfonts, amsthm, mathrsfs}
\usepackage[title]{appendix}
\usepackage{xcolor}
\usepackage{textcomp}
\usepackage{lmodern}
\usepackage{manyfoot}
\usepackage{booktabs}
\usepackage{algorithm}
\usepackage{algorithmicx}
\usepackage{algpseudocode}
\usepackage{listings}

\usepackage{dsfont}
\usepackage{makecell}
\usepackage{soul}
\usepackage{subcaption}

\begin{document}

\title[A Practical Approach to Novel Class Discovery in Tabular Data]{A Practical Approach to Novel Class Discovery in Tabular Data}

\author*[1,2]{\fnm{Troisemaine} \sur{Colin}}\email{colin.troisemaine@gmail.com}
\author[1]{\fnm{Reiffers-Masson} \sur{Alexandre}}
\author[2]{\fnm{Gosselin} \sur{St\'ephane}}
\author[2]{\fnm{Lemaire} \sur{Vincent}}
\author[1]{\fnm{Vaton} \sur{Sandrine}}

\affil[1]{\orgdiv{Department of Computer Science}, \orgname{IMT Atlantique}, \orgaddress{\city{Brest}, \country{France}}}

\affil[2]{\orgname{Orange Labs}, \orgaddress{\city{Lannion}, \country{France}}}

\abstract{
The problem of Novel Class Discovery (NCD) consists in extracting knowledge from a labeled set of known classes to accurately partition an unlabeled set of novel classes.
While NCD has recently received a lot of attention from the community, it is often solved on computer vision problems and under unrealistic conditions.
In particular, the number of novel classes is usually assumed to be known in advance, and their labels are sometimes used to tune hyperparameters.
Methods that rely on these assumptions are not applicable in real-world scenarios.
In this work, we focus on solving NCD in tabular data when no prior knowledge of the novel classes is available.
To this end, we propose to tune the hyperparameters of NCD methods by adapting the $k$-fold cross-validation process and hiding some of the known classes in each fold.
Since we have found that methods with too many hyperparameters are likely to overfit these hidden classes, we define a simple deep NCD model.
This method is composed of only the essential elements necessary for the NCD problem and shows robust performance under realistic conditions.
Furthermore, we find that the latent space of this method can be used to reliably estimate the number of novel classes.
Additionally, we adapt two unsupervised clustering algorithms ($k$-means and Spectral Clustering) to leverage the knowledge of the known classes.
Extensive experiments are conducted on 7 tabular datasets and demonstrate the effectiveness of the proposed method and hyperparameter tuning process, and show that the NCD problem can be solved without relying on knowledge from the novel classes.
}


\maketitle

\section{Introduction}
Recently, remarkable progress has been achieved in supervised tasks, in part with the help of large and fully labeled sets such as ImageNet \cite{deng2009imagenet}.
These advancements have predominantly focused on closed-world scenarios, where, during training, it is presumed that all classes are known in advance and have some labeled examples.
However, in practical applications, obtaining labeled instances for all classes of interest can be a difficult task due to factors such as budget constraints or lack of comprehensive information.
Furthermore, for models to be able to transfer learned concepts to new classes, they need to be designed with this in mind from the start, which is rarely the case.
Yet this is an important skill that humans can use effortlessly.
For example, having learnt to distinguish a few animals, a person will easily be able to recognise and ``cluster'' new species they have never seen before.
The transposition of this human capacity to the field of machine learning could be a model capable of categorizing new products in novel categories.

This observation has led researchers to formulate a new problem called Novel Class Discovery (NCD) \cite{tr2023introduction, hsu2018learning}.
Here, we are given a labeled set of known classes and an unlabeled set of different but related classes that must be discovered.
Lately, this task has received a lot of attention from the community, with many new methods such as AutoNovel \cite{autonovel2021}, OpenMix \cite{zhong2020openmix} or NCL \cite{zhong2021neighborhood} and theoretical studies \cite{sun2023and, li2022a}.
However, most of these works tackle the NCD problem under the unrealistic assumption that the number of novel classes is known in advance, or that the target labels of the novel classes are available for hyperparameter optimization \cite{tr2022method}.
These assumptions render these methods impractical for real-world NCD scenarios.
To address these challenges, we propose a general framework for optimizing the hyperparameters of NCD methods where the ground-truth labels of novel classes are never used, as they are not available in real-world NCD scenarios.
Furthermore, we show that the latent spaces obtained by such methods can be used to accurately estimate the number of novel classes.

We also introduce three new NCD methods.
Two of them are unsupervised clustering algorithms modified to leverage the additional information available in the NCD setting.
The first one improves the centroid initialization step of $k$-means, resulting in a fast and easy to use algorithm that can still give good results in many scenarios.
The second method focuses on optimizing the parameters of the Spectral Clustering (SC) algorithm.
This approach has a potentially higher learning capacity as the representation itself (i.e. the spectral embedding) is tuned to easily cluster the novel data.
Finally, the last approach is a deep NCD method composed of only the essential components necessary for the NCD problem.
Compared to SC, this method is more flexible in the definition of its latent space and effectively integrates the knowledge of the known classes.

While these contributions can be applied to any type of data, our work focuses on tabular data.
The NCD community has focused almost exclusively on computer vision problems and, to the best of our knowledge, only one paper \cite{tr2022method} has tackled the problem of NCD in the tabular context.
This is despite the many applications of NCD in tabular data.
Some examples include \textit{cybersecurity}, where network traffic could be used to infer new types of cyber-attacks that deviate from known attack patterns; \textit{sensor data analysis}, where new categories of equipment failures could be diagnosed and lead to faster repairs; or \textit{fraud detection}, where new types of fraudulent transactions that deviate from known patterns could be discovered.
However, this previous work \cite{tr2022method} required the meticulous tuning of a large number of hyperparameters to achieve optimal results.
Methods designed for tabular data cannot take advantage of powerful techniques commonly employed in computer vision.
Examples include convolutions, data augmentation or Self-Supervised Learning methods such as DINO \cite{dino}, which have been used with great success in NCD works \cite{vaze2022generalized, fei2022xcon, zhang2022automatically}, thanks to their strong ability to obtain representative latent spaces without any supervision.
On the other hand, tabular data methods have to rely on finely tuned hyperparameters to achieve optimal results.
For this reason, we believe that the field of tabular data will benefit the most from our contributions.

By making the following contributions, we demonstrate the feasibility of solving the NCD problem with tabular data and under realistic conditions:
\begin{itemize}
    \item We develop a hyperparameter optimization procedure tailored to transfer the results from the known classes to the novel classes with good generalization.
    \item We show that it is possible to accurately estimate the number of novel classes in the context of NCD, by applying simple clustering quality metrics in the latent space of NCD methods.
    \item We modify two classical unsupervised clustering algorithms to effectively utilize the data available in the NCD setting.
    \item We propose a simple and robust method, called PBN (for \textit{Projection-Based NCD}), that learns a latent representation that incorporates the important features of the known classes, without overfitting on them.
\end{itemize}

\noindent The code is available at \hbox{\url{https://github.com/Orange-OpenSource/PracticalNCD}}.

\section{Related work}
The setup of NCD \cite{tr2023introduction}, which involves both labeled and unlabeled data, can make it difficult to distinguish from the many other domains that revolve around similar concepts.
In this section, we review some of the most closely related domains and try to highlight their key differences in order to provide the reader with a clear and comprehensive understanding of the NCD domain.

\textbf{Semi-supervised Learning} is another domain that is at the frontier between supervised and unsupervised learning.
Specifically, a labeled set is given alongside an unlabeled set containing instances that are assumed to be from the same classes.
Semi-supervised Learning can be particularly useful when labeled data is scarce or annotation is expensive.
As unlabeled data is generally available in large quantities, the goal is to exploit it to obtain the best possible generalization performance given limited labeled data.

The main difference with NCD is that the all the classes are known in advance.
Some works have shown that the presence of novel classes in the unlabeled set negatively impacts the performance of Semi-Supervised Learning models \cite{chen2020semi, guo2020safe}.
So as these works do not attempt to discover the novel classes, they are not applicable to NCD.

\textbf{Transfer Learning} aims at solving a problem faster or with better performance by leveraging knowledge from a different problem.
It is commonly expressed in computer vision by pre-training models on ImageNet \cite{deng2009imagenet}.
Transfer Learning can be either \textit{cross-domain}, when a model trained on a given dataset is fine-tuned to perform the same task on a different (but related) dataset.
Or it can be \textit{cross-task}, where a model that can distinguish some classes is re-trained for other classes of the same domain.

NCD can be viewed as a cross-task Transfer Learning problem where the knowledge from a classification task on a source dataset is transferred to a clustering task on a target dataset.
But unlike NCD, Transfer Learning typically requires the target spaces of both sets to be known in advance.
Initially, NCD was characterized as a Transfer Learning problem (e.g. in DTC \cite{han2019learning} and MCL \cite{hsu2019multiclass}) and the training was done in two stages: first on the labeled set and then on the unlabeled set.
This methodology seemed natural, as with Transfer Learning, both sets are not available at the same time.

The field of \textbf{Open Set Recognition} (OSR) \hbox{\cite{Scheirer2013}} stems from the observation that neural networks have a tendency to make predictions with high confidence, even for instances of classes they have never seen before.
It is based on the Open-World assumption in which instances of new classes may appear during inference, so standard models cannot be used.
Therefore, in OSR, samples from novel classes must be correctly identified as unknown.
An example would be a fingerprint recognition model that needs to reject images of fingerprints (or anything else!) from people who aren't registered.
However, as in many domains under the Open-World assumption, the goal of OSR is only to identify samples from novel classes.
In NCD, we assume that this identification has already been made, and we try to actually discover the novel classes.

\textbf{Generalized Category Discovery} (GCD) was first introduced by \cite{vaze2022generalized} and has also attracted attention from the community \cite{zheng2022openset, Yang_2022_CVPR, fei2022xcon}.
It can be seen as a less constrained alternative to NCD, since it does not rely on the assumption that samples belong exclusively to the novel classes during inference.
However, this is a more difficult problem, as the models must not only cluster the novel classes, but also accurately differentiate between known and novel classes while correctly classifying samples from the known classes.

Some notable works in this area include ORCA \cite{cao2022openworld} and OpenCon \cite{sun2023opencon}.
Namely, ORCA trains a discriminative representation by balancing a supervised loss on the known classes and unsupervised pairwise loss on the unlabeled data.
And OpenCon proposes a contrastive learning framework which employs Out-Of-Distribution strategies to separate known vs. novel classes.
Its clustering strategy is based on moving prototypes that enable the definition of positive and negative pairs of instances.

\textbf{Novel Class Discovery} has a rich body of papers in the domain of computer vision.
Early works approached this problem in a \textit{two-stage} manner.
Some define a latent space using only the known classes, and project the unlabeled data into it (DTC\cite{han2019learning} and MM \cite{chi2022meta}).
Others train a pairwise labeling model on the known classes and use it to label and then cluster the novel classes (CCN \cite{hsu2018learning} and MCL \cite{hsu2019multiclass}).
But both of these approaches suffered from overfitting on the known data when the high-level features were not fully shared by the known and novel classes.

Today, to alleviate this overfitting, the majority of approaches are \textit{one-stage} and try to transfer knowledge from labeled to unlabeled data by learning a shared representation.
In this category, AutoNovel \cite{autonovel2021} is one of the most highly influential works.
After pre-training their latent representation with SSL \cite{gidaris2018unsupervised}, two classification networks are jointly trained.
The first simply learns to distinguish the known classes with the ground-truth labels.
And the other learns to separate unlabeled data from pseudo-labels defined for each epoch based on pairwise similarity.
NCL \cite{zhong2021neighborhood} adopts the same architecture as AutoNovel, and extends the loss by adding a contrastive learning term to encourage the separation of novel classes.
OpenMix \cite{zhong2020openmix} utilizes the MixUp strategy to generate more robust pseudo-labels.

As expressed before, although these methods have achieved some success, they are not applicable to tabular data.
To date, and to the best of our knowledge, only TabularNCD \cite{tr2022method} tackles this problem.
Also inspired by AutoNovel, it pre-trains a dense-layer autoencoder with SSL and adopts the same loss terms and dual classifier architecture.
Pseudo-labels are defined between pairs of unlabeled instances by checking if they are among the most similar pairs.

For a more complete overview of the state-of-the-art of NCD, we refer the reader to the survey \cite{tr2023introduction}.

\section{Approaches}

In this section, after introducing the notations, we define two simple but potentially effective models derived from classical clustering algorithms (Sections \ref{sec:ncdkmeans} and \ref{sec:ncdsc}).
The idea is to use the labeled data to improve the unsupervised clustering process, and make the comparison to NCD methods more challenging.
Then, we present a new method, PBN (for \ul{P}rojection-\ul{B}ased \ul{N}CD, Section \ref{sec:pbn}), characterized by its low number of hyperparameters needed to be tuned.

\subsection{Problem setting}
We start by describing the Novel Class Discovery setup and the necessary notations.
Here, data is provided in two distinct sets:
a labeled set of known classes $D^l = \left\{ \left( x_i^l, y_i^l \right) \right\}^N_{i=1}$ with $x_i^l \in \mathcal{X} = \mathbb{R}^d$ and $y_i^l \in \mathcal{Y}^l = \left\{ 1, \dots, C^l \right\}$ the ground-truth labels of $x_i^l$.
And an unlabeled set $D^u = \left\{ \left( x_i^u \right) \right\}^M_{i=1}$ where the data samples $x_i^u \in \mathcal{X}$ are only from novel classes $\mathcal{Y}^u = \left\{ 1, \dots, C^u \right\}$, which are different but related to the known classes. 
In other words, there is no overlap between the known and novel classes, so $\mathcal{Y}^l \cap \mathcal{Y}^u = \emptyset$.
The objective is to exploit the knowledge from $D^l$ to accurately partition $D^u$ into the $C^u$ clusters of the novel classes.

Following previous research, we first assume that the number of novel classes $C^u$ is known in advance, and later propose an approach to estimate this number in a later section.

\subsection{NCD k-means}
\label{sec:ncdkmeans}
This is a straightforward method that takes inspiration from $k$-means++ \cite{arthur2007k}, which is an algorithm for choosing the initial positions of the centroids (or cluster centers).
In $k$-means++, the first centroid is chosen at random from the data points.
Then, each new centroid is chosen iteratively from one of the data points with a probability proportional to the squared distance from the point's closest centroid.
The resulting initial positions of the centroids are generally spread more evenly, which yields appreciable improvement in the final error of k-means and convergence time.

\begin{figure}[htbp]
    \hspace*{\fill}
    \begin{subfigure}{0.35\textwidth}
        \includegraphics[width=1\textwidth]{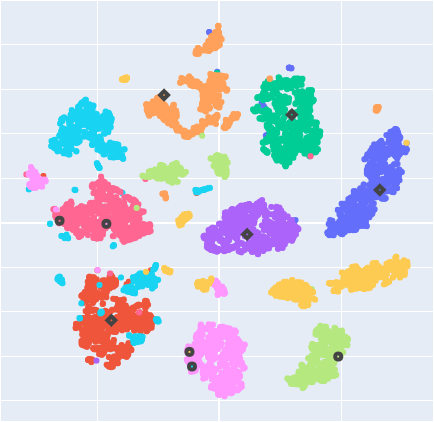}
        \caption{Before convergence.}
        \label{fig:cent1}
    \end{subfigure}
    \hfill
    \begin{subfigure}{0.35\textwidth}
        \includegraphics[width=1\textwidth]{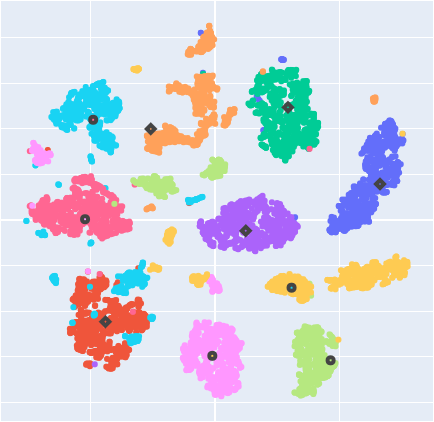}
        \caption{After convergence.}
        \label{fig:cent2}
    \end{subfigure}
    \hspace*{\fill}
    \caption{t-SNE plots of the \textit{Pendigits} dataset depicting the centroids before and after convergence. Note how the centroids of the known classes (the squares) don't move, as they stay the mean class point.}
    \label{fig:centroids_before_after_convergence}
\end{figure}

As shown in Figure~\ref{fig:cent1}, we naively adapt $k$-means++ to the NCD setting by defining $C^l$ initial centroids.
They are set as the mean class points of the known classes using the ground-truth labels.
Then, we follow $k$-means++ and randomly select $C^u$ new centroids in the unlabeled set, with similarly decreasing probability when closer to existing centroids.
We found experimentally (see Appendix~\ref{sec:appendix_centroids}) that, after the initialization is complete, the best accuracy is achieved when only the centroids of the novel classes are updated, and using the unlabeled data only.
In other words, the data of the known classes is only used during the initialization of the new centroids, but not during the convergence phase.
Intuitively, if the centroids of the known and novel data are updated together, they have a higher risk of drifting and capturing data of the other set.
The pseudocode of the proposed method is summarized in Algorithm~\hbox{\ref{alg:ncdkmeans}} of Appendix~\hbox{\ref{sec:appendix_pseudocode}}.

Similarly to $k$-means++, we repeat the initialization process a few times and keep the centroids that achieved the smallest inertia.
Note that, to stay consistent with the $k$-means algorithm, we use also the $L_2$ norm (i.e. the Euclidean distance) for NCD $k$-means.

\subsection{NCD Spectral Clustering}
\label{sec:ncdsc}
Spectral Clustering (SC) is an alternative to distance-based clustering methods (such as $k$-means).
It makes no assumptions about the structure of the data and considers the clustering problem as a graph partitioning problem and seeks to decompose the graph into connected components \cite{Luxburg07}.
The input of the Spectral Clustering algorithm is an adjacency matrix (sometimes called similarity graph) which must accurately represent the neighborhood relationships between data points.
There are multiple ways to construct such a graph, however there is no theoretical result on the relation between the similarity graph construction method and the Spectral Clustering results.
Here, we employ a popular approach to construct the adjacency matrix, which is through a Gaussian kernel: $A_{i,j} = \exp \left({-\lVert x_i - x_j \rVert^2_2 / (2\sigma^2)} \right), \forall x_i,x_j \in \mathcal{X}$ where the parameter $\sigma$ controls the width of the neighborhood.

Following the Ng-Jordan-Weiss algorithm~\cite{NIPS2001_801272ee}, we use the symmetric normalized Laplacian $L_{sym} = D^{-1/2}LD^{-1/2}$, where $D$ is the degree matrix defined as: $D_{i,i} = \sum_{j=1}^{n}A_{i,j}$, in which $n$ is the number of data samples.
The next step consists in finding the first $u$ eigenvectors of $L_{sym}$ to form the spectral embedding $U \in \mathbb{R}^{n \times u}$, where row $i$ corresponds to point $x_i$.
Finally, the points in $U$ are partitioned with $k$-means into clusters.

The optimal value of $\sigma$ in the Gaussian kernel can vary widely depending on the distribution of inter-point distances.
For this reason, we take inspiration from the rules of thumb given in \cite{Luxburg07} and employ a minimum spanning tree (MST) to choose $\sigma$.
In the past few years, several graph-based clustering methods that use the MST have been proposed \cite{khan2022fast}, as it reliably represents the layout of the data and is inexpensive to compute.
In the approach proposed here, we denote $d_{max}$ the length of the longest edge in the MST of inter-point distances.
The longest edge of the MST is a much studied object \cite{stuetzle2003estimating} and is representative of the scale of the dataset.
The scaling factor $\sigma$ is then calculated such that, after applying the Gaussian kernel, $d_{max}$ is transformed into a chosen similarity $s_{min}$.
This ensures that the resulting graph is safely connected. By optimizing this similarity $s_{min}$ we can accurately represent the neighborhood relationships.

Therefore, for a given value of $s_{min}$, we derive $\sigma$ from the length of the longest edge $d_{max}$ in the MST:
\begin{align}
    \exp \left( - {d^2_{max}} / (2\sigma^2) \right) &= s_{min} \nonumber \\
    \Leftrightarrow \sigma &= d_{max} / \sqrt{ - 2 \ln \left( s_{min} \right)}
    \label{eq:thumb}
\end{align}
Optimizing $s_{min}$ instead of $\sigma$ should be more robust to variations in the distribution of inter-point distances and give better results across different datasets or parts of a dataset.

To incorporate the knowledge from the known classes in the Spectral Clustering process, our initial approach was to utilize NCD $k$-means within the spectral embedding.
In short, we would initially compute the full spectral embedding for all the data and determine the mean points of the known classes with the help of the ground truth labels.
These mean points would then serve as the initial centroids.
However, the observed performance improvement over the fully unsupervised SC was quite marginal.

Instead, the idea that we will use throughout this article (and that we refer to as ``NCD Spectral Clustering'') stems from the observation that SC can obtain very different results according to the parameters that are used.
Among these parameters, the temperature parameter $\sigma$ of the kernel holds particular importance, as it directly impacts the adjacency matrix's accuracy in representing the neighborhood relationships of the data points.
The rule of thumb of Equation~\ref{eq:thumb} still requires to choose a value, but significantly reduces the space of possible values.
Additionally, while the literature often sets the number of components $u$ of the spectral embedding equal to the number of clusters, we have observed that optimizing it can also improve performance.

\begin{figure}[htbp]
	\centering
    \includegraphics[width=0.65\textwidth]{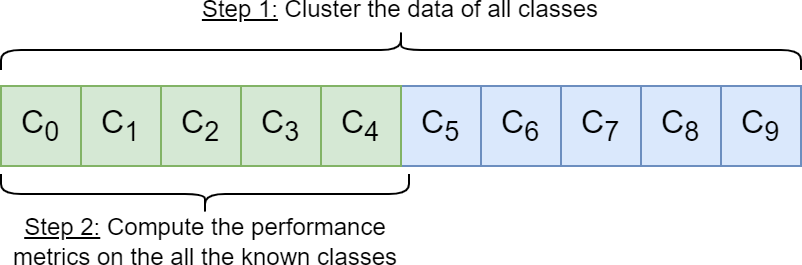}
    \caption{NCD Spectral Clustering parameter optimization process.}
    \label{fig:ncd_sc}
\end{figure}

Therefore, rather than a specific method, we propose the parameter optimization scheme illustrated in Figure~\ref{fig:ncd_sc}.
For a given combination of parameters $\{ s_{min}, u \}$, the corresponding spectral embedding of all the data is computed and then partitioned with $k$-means.
The quality of the parameters is evaluated from the clustering performance on the known classes, as the ground-truth labels are only available for the known classes.
Indeed an important hypothesis behind the NCD setup is that the known and novel classes are related and share similarities, so they should have similar feature scales and distributions.
Consequently, if the Spectral Clustering performs well on the known classes, the parameters are likely suitable to represent the novel classes.
The pseudocode of this approach can be found in Algorithm~\hbox{\ref{alg:ncdsc}} of Appendix~\hbox{\ref{sec:appendix_pseudocode}}.

\textbf{Discussion.}
This idea can be applied to optimize the parameters of any unsupervised clustering algorithm in the NCD context.
For example, the $Eps$ and $MinPts$ parameters of DBSCAN \cite{ester1996density} can be selected in the same manner.
It is also possible to use a different adjacency matrix in the SC algorithm.
One option could be to substitute the Gaussian kernel with the $k$-nearest neighbor graph, and therefore optimise $k$ instead of $\sigma$.
However, for the sake of simplicity, we will only investigate SC using the Gaussian kernel.

\subsection{Projection-Based NCD}
\label{sec:pbn}

\begin{figure}[htbp]
	\centering
    \includegraphics[width=\textwidth]{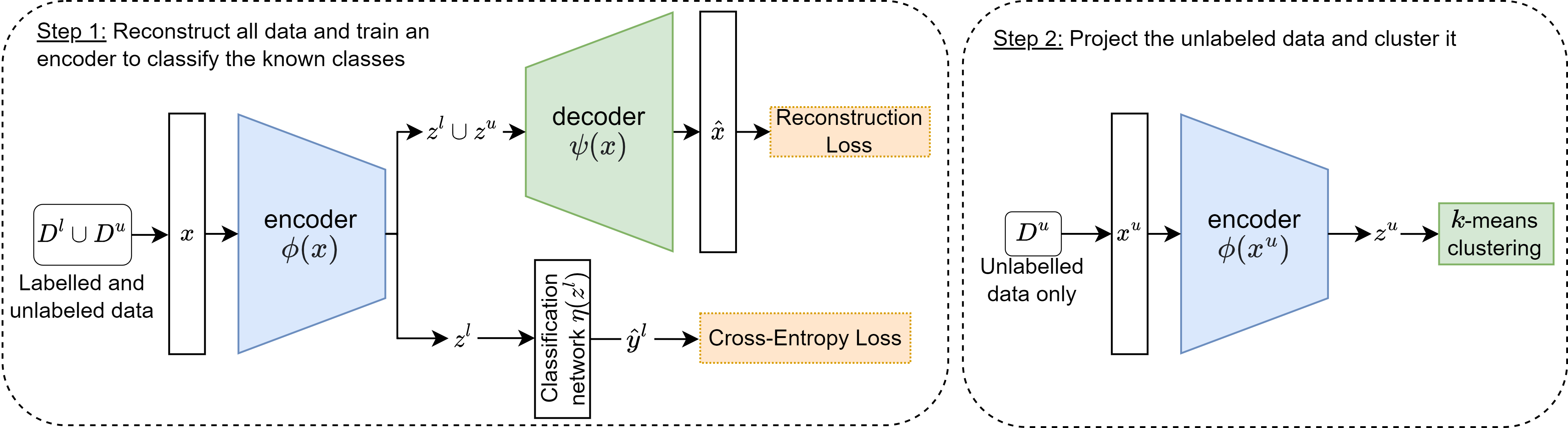}
    \caption{Architecture of the PBN model.}
    \label{fig:projectionmodel}
\end{figure}

Projection-based NCD (PBN) can be seen as an extension of the baseline method used in TabularNCD \cite{tr2022method}.
PBN is illustrated in Figure~\ref{fig:projectionmodel} and consists of 3 key components: (1) an encoder that learns a shared representation between the known and novel classes; (2) a classification network trained to distinguish the known classes of $D^l$ in order to incorporate their relevant features into the representation; and (3) a decoder that reconstructs the data for both known and novel classes of $D^l \cup D^u$, ensuring that the latent space contains the information necessary to represent all classes.
The decoder serves a dual purpose: it provides regularization and mitigates overfitting on the known classes, thus improving generalization, as shown in \cite{le2018supervised}.

The training loss is defined as:
\begin{equation}
    \mathcal{L}_{PBN} = w \times \mathcal{L}_{CE} + (1 - w) \times \mathcal{L}_{MSE}
    \label{eq:ProjecNCDLoss}
\end{equation}
where $w \in (0,1)$ is a trade-off parameter that allows to balance the strength of the cross-entropy loss and the reconstruction loss.

The cross-entropy loss on the known classes is defined as:
\begin{equation}
     \mathcal{L}_{CE} = - \sum_{c=1}^{C^l} y_c \log \left( 
\eta_{c}(z)
 \right)
\end{equation}
where $\eta(z) = \left( \eta_{c}(z) \right)^{C^l}_{c=1}$ is the output of a classification network composed of a single dense layer of neurons, $z = \phi(x)$ is the projection of instance $x$ through the encoder $\phi$ and $\left( y_{c} \right)^{C^l}_{c=1}$ is the one-hot encoded ground-truth label of instance $x$.

The reconstruction loss of the instances from all classes is written as:
\begin{equation}
    \mathcal{L}_{MSE} = \frac{1}{d} \sum_{j=1}^d \left( x_j - \hat{x}_j \right)^2
\end{equation}
where $\hat{x} = \psi(z)$ is the reconstruction of instance $x \in \mathbb{R}^d$.

Once the encoder, decoder and classification network have been trained (step 1), unlabeled data $D^u$ is projected by the trained encoder into the latent space and then clustered with $k$-means to discover novel classes (step 2).

Projection-based NCD requires tuning of four hyperparameters.
The trade-off parameter $w$ is inherent to the method and the other three come from the choice of architecture: the learning rate, the dropout rate and the size of the latent space.

Note that this method doesn't employ complex schemes to define pseudo-labels unlike many NCD works.
They have been proven to be accurate with image data (notably thanks to data augmentation techniques) \cite{autonovel2021, zhao2021novel}, but we found in preliminary results not detailed here, that for tabular data, they introduce variability in the results and new hyperparameters that need to be tuned.

\textbf{Discussion.}
Similarly to PBN, the baseline method of TabularNCD \cite{tr2022method} relies on the assumption that known and novel classes share similar high-level features, and defines a latent space that highlights these features.
This baseline first trains a deep classifier to distinguish only the known classes of $D^l$.
After training, the output and softmax layers are discarded, and the last hidden layer is now considered as the output of an encoder.
It then projects the novel data of $D^u$ into this latent space and partitions it using $k$-means.
This is the basic workflow of two-stage latent space-based NCD methods identified in \cite{tr2023introduction}.
It is also similar to DTC \cite{han2019learning}, which uses the more refined DEC \cite{xie2016unsupervised} clustering model instead of $k$-means in the baseline.
The problem with such two-stage methods is that the resulting representations are at risk of being heavily biased towards the known classes.
Thus, if some concepts or high-level features are not shared between the known and novel classes, the novel classes will not be well represented and these approaches will fail.

\subsection{Summary of proposed approaches}
In this section, we have proposed 3 distinct methods for solving the NCD problem, all of which leverage knowledge from the known classes in different ways.
Firstly, NCD $k$-means uses the labeled data to improve the initialization of its centroids.
Secondly, instead of using the labels of the known classes during the clustering process itself, NCD Spectral Clustering uses them to find parameters that are likely to be suitable for the whole domain.
More precisely, by clustering $D^l \cup D^u$ together, the adequacy of the parameters $s_{min}$ and $u$ can be evaluated on the known classes.
Finally, PBN is a straightforward method that includes only the essential components to define a latent representation suitable for clustering the novel classes.
In this case, an encoder is trained with a classification loss on the known classes and a reconstruction loss on all the data to ensure that the novel classes are not misrepresented.
The novel data is then projected into this representation and clustered with $k$-means.

In the next section, we present an approach to finding hyperparameters without using the labels of the novel classes, which are not available in realistic scenarios. Indeed in the experiments (see Section~\ref{sec:exp}), it should become clear why the simplicity of the proposed approach is a desirable feature for hyperparameter optimization in the NCD context.

\section{Hyperparameter optimization}
\label{sec:hp}
The success of machine learning algorithms (including NCD) can be attributed in part to the high flexibility induced by their hyperparameters.
In most cases, a target is available and approaches such as the $k$-fold Cross-Validation (CV) can be employed to tune the hyperparameters and achieve optimal results.
However, in a realistic scenario of Novel Class Discovery, the labels of the novel classes are never available.
We must therefore find a way to optimize hyperparameters without ever relying on the labels of the novel classes.
In this section, we present a method that leverages the known classes to find hyperparameters applicable to the novel classes.
This tuning method is designed specifically for NCD algorithms that require both labeled data (known classes) and unlabeled data (novel classes) during training\footnote{To optimize purely unsupervised clustering methods for NCD, we refer the reader to the optimization process of Section~\ref{sec:ncdsc}.}.
This is the case for Projection-based NCD, as described in Section ~\ref{sec:pbn}.

\begin{figure}[htbp]
	\centerline{\includegraphics[width=0.6\textwidth]{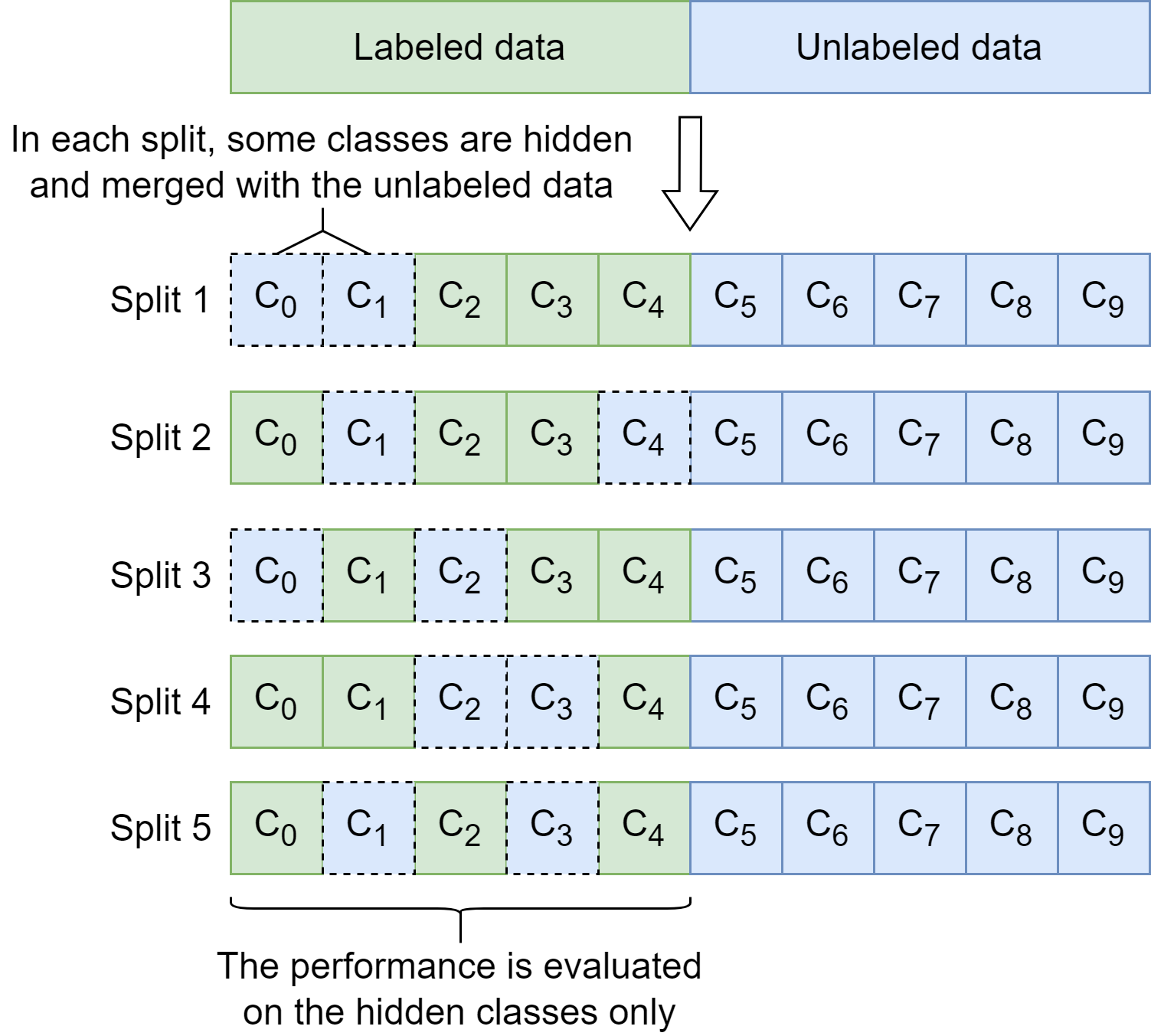}}
    \caption{The $k$-fold cross-validation approach for hyperparameter optimisation of NCD methods.}
    \label{fig:kfold}
\end{figure}

The process that we devised is represented in Figure~\ref{fig:kfold}.
For each of the splits, the instances of around half of the known classes are selected to form the set $D^{hid}$ and their labels are hidden.
The labeled set now becomes the instances of $D^l \setminus D^{hid}$ and the unlabeled set becomes the instances of $D^u \cup D^{hid}$.
After training the model with this new data split, it is evaluated for its performance for partitioning the instances of $D^{hid}$ only since their labels are available.

To illustrate, in the split 1 of Figure~\ref{fig:kfold}, the model will be trained with the subsets of classes $\left\{ C_2, C_3, C_4 \right\}$ as known classes and $\left\{ C_0, C_1, C_5, \dots, C_9 \right\}$ as novel classes.
It will be evaluated for its performance on the hidden classes $\left\{ C_0, C_1 \right\}$ only.

To evaluate a given combination of hyperparameters, this approach is applied to all the splits, and the performance on the hidden classes is averaged.
After repeating this process for many combinations, the combination that achieved the best performance is selected.
For the final evaluation on the novel classes, in a realistic scenario of NCD their labels are never available.
However, in the datasets employed in this article, the novel classes are comprised of pre-defined classes.
Therefore, even though these labels are not employed during training, they can still be used to assess the final performance on the novel classes of different models and compare them against each other.

This tuning method stems from the same idea behind the NCD Spectral Clustering parameterization process.
Namely, if the clustering in the learned representation successfully partitions the hidden classes in $D^{hid}$, it is also likely suitable for the novel classes in $D^u$.
Furthermore, keeping the unlabeled data during training even though the model is not evaluated on the novel classes is important, as it increases the chances of the representation being adapted for all the classes.
For the same reason, the $k$-means of PBN (see Section~\ref{sec:pbn}) is fitted on $D^u \cup D^{hid}$ together (instead of just $D^{hid}$) and the performance is then computed on $D^{hid}$ only.
So cases where the classes in $D^u$ and $D^{hid}$ are tangled will be penalized.

In Table~\ref{table:kfold_splits}, we report for all datasets used in our experiments the number of known classes that are hidden in each split, as well as the number of splits.
Note that when the number of known classes is small (e.g. 3 for \textit{Human}), this approach may be difficult to apply.

\begin{table}[htbp]
    \caption{Classes splits of the $k$-fold cross-validation.}
    \label{table:kfold_splits}
    \centering
    \begin{tabular}{l | c c | c c}
        \hline
        \multirow{2}{*}{Dataset} & Known & Novel & Hidden & \multirow{2}{*}{Splits} \\
         & classes & classes & classes & \\
        \hline
        Human       & 3  & 3  & 2  & 3 \\
        Letter      & 19 & 7  & 7  & 5 \\
        Pendigits   & 5  & 5  & 2  & 5 \\
        Census      & 12 & 6  & 6  & 5 \\
        m feat      & 5  & 5  & 2  & 5 \\
        Optdigits   & 5  & 5  & 2  & 5 \\
        CNAE-9      & 4  & 5  & 2  & 5 \\
        \hline
    \end{tabular}
\end{table}

\textbf{Discussion.}
Similarly to NCD, there are no labels available in unsupervised clustering problems, which makes the task of hyperparameter selection very difficult.
To address this issue, clustering algorithms are sometimes tuned using \textit{internal} metrics that do not rely on labeled data for computation.
These metrics offer a means of comparing the results obtained from different clustering approaches.
Examples of such metrics include the Silhouette coefficient, Davies-Bouldin index, or Calinski-Harabasz index \cite{arbelaitz2013extensive}.
However, it is important to note that these metrics make assumptions about the structure of the data and can be biased towards algorithms which make a similar assumption.
But unlike unsupervised clustering, the NCD setting provides known classes that are related to the novel classes we are trying to cluster.

\section{Estimating the number of novel classes}
\label{sec:est}
Cluster Validity Indices (CVIs) are commonly used in unsupervised data analysis to estimate the number of clusters and are also applicable to the NCD problem.
CVIs are scores that compare the compactness and separation of clusters without the help of external information such as ground truth labels.
However, the knowledge from the known classes isn't used if the CVIs are directly applied to estimate the number of novel classes.
Therefore, we propose to apply the CVIs in the latent representation learned by PBN.
Projection-based NCD methods such as PBN are designed to create a latent space that emphasizes the relevant features of the known classes.
Since these features are shared to some extent with the novel classes, this representation should be better at revealing the clusters we are trying to discover than the original feature space.
Consequently, it makes sense that applying the different estimation techniques in the learned latent space should yield better results.

Note that this is only applicable to NCD methods such as PBN that don't require the number of novel classes $C^u$ to train their latent space (unlike TabularNCD).
For the others, the estimation can be done once in the original feature space, but should have higher error.
~\\
\par
Some NCD works have also previously attempted to estimate the number of novel classes.
For instance, \cite{hsu2018learning} defines a large number of output neurons in their clustering network (e.g. 100).
In this case, the clustering network is expected to use only the necessary number of clusters while leaving the remaining output neurons unused.
Clusters were counted if they contained more instances than a certain threshold.
However, since, with the exception of TabularNCD, the models studied in this paper do not use a clustering network, we will not evaluate this method.

Another technique, proposed by \cite{vaze2022generalized}, consists in training a $k$-means on the combined dataset $D^l \cup D^u$ and selecting the $k$ that yielded the highest accuracy on $D^l$. 
While this approach worked well for balanced datasets \cite{fei2022xcon}, it has been shown to underperform in the case of unbalanced class distributions \cite{yang2023bootstrap}.
For the sake of simplicity, we will call this method KM-ACC (for \ul{$k$}-\ul{m}eans \ul{ACC}) in the remainder of this paper.
~\\
\par
To select the CVI that we will use for our application, we rely on the results of \cite{arbelaitz2013extensive}.
Here, the authors conducted an extensive performance evaluation of 30 CVIs.
They concluded that the Silhouette, Davies–Bouldin, Calinski–Harabasz and Dunn indices behaved better than other indices in almost all cases.
In the experiments, the performance of these 4 indices will be compared, with the addition of the elbow method and the NCD-specific method KM-ACC.

\section{Full training procedure}
\label{sec:full_proce}
In the previous sections, we presented the models, the hyperparameter optimization and the estimation procedure of the number of novel classes independently.
In this section, these components are brought together to form a complete training procedure.
To ensure that no prior knowledge about the novel classes is ever used in this process, the number of novel classes is naturally estimated during the $k$-fold CV introduced in Section~\ref{sec:hp}.
As the whole process is quite complex, we try to summarize it in clear terms in this section and in Algorithm \ref{alg:cap}.

To gauge a given set of hyperparameters, we evaluate the performance of the model over $n_{folds}$, where in each fold, a random combination of known classes is ``hidden'' and merged with the unlabeled data of $D^u$.
In a fold, the encoder of the NCD model is first trained on this new data split. 
The number of novel classes is then estimated with a CVI in the projection of the unlabeled data.
At this point, the novel and hidden classes are partitioned by the model in the latent space using the previous estimate of the number of clusters.
And the accuracy for this fold is calculated on the hidden classes.
This process is repeated for all folds and for many combinations of hyperparameters, and the combination that achieved the best performance on average is selected for the final evaluation of the model.

\begin{algorithm}
    \caption{Agnostic NCD model evaluation}
    \label{alg:cap}
    \begin{algorithmic}[1]
        \Require Training data $\{ D^l, D^u \}$, hyperparameters $\theta$, number of classes to hide $n_{hid}$, number of folds $n_{folds}$
        \State \textbf{Initialize:} $folds \gets$ set of $n_{folds}$ random combinations of $n_{hid}$ known classes
        \For{each $fold$ \textbf{in} $folds$}
            \State $D^{hid} \gets$ the data from $D^l$ of the classes in $fold$
            \State $D^{l^{\prime}} \gets D^l \setminus D^{hid}$
            \State $D^{u^{\prime}} \gets D^u \cup D^{hid}$
            \State Train model on $\{ D^{l^{\prime}}, D^{u^{\prime}} \}$ with hyperparameters $\theta$
            \State $Z^u \gets \phi_{\theta}(X^u)$ the projection of the novel data
            \State $k^{\prime} \gets$ the estimation of $C^u$ in $Z^u$ with a CVI
            \State Get the clustering prediction of the model for $D^{u^{\prime}}$ using $k = n_{hid} + k^{\prime}$
            \State $ACC_{hid} \gets$ clustering performance on $D^{hid}$
        \EndFor
        \State \textbf{Return:} Average of all the $ACC_{hid}$
    \end{algorithmic}
\end{algorithm}

\section{Experiments}
\label{sec:exp}

\subsection{Experimental setup}
~\par
\textbf{Datasets.}
To evaluate the performance of the methods compared in this paper, 7 tabular classification datasets were selected: Human Activity Recognition \cite{humanactivity}, Letter Recognition \cite{letterrecog}, Pen-Based Handwritten Digits \cite{uci}, 1990 US Census Data \cite{uci}, Multiple Features \cite{uci}, Handwritten Digits \cite{uci} and CNAE-9 \cite{uci}. 

Following the previous NCD works \cite{zhong2021neighborhood, han2019learning, autonovel2021}, the instances of about 50\% of the classes are hidden a priori to form the unlabeled set of novel classes $D^u$, while the rest form the labeled set $D^l$.
We use a 70/30\% train/test split if it was not already provided.
Statistical information on the datasets is shown in Table~\ref{table:data}, and the number of known/novel classes (along with the number of classes hidden during the $k$-fold CV) can be found in Table~\ref{table:kfold_splits}.
The numerical features of all the datasets are pre-processed to have zero mean and unit variance, while the categorical features are one-hot encoded.

\begin{table}[htbp]
    \caption{Details of the datasets.}
    \label{table:data}
    \centering
        \begin{tabular}{| l | c | c | c | c | c | c | c |}
            \hline
            Dataset       & Human & Letter & Pendigits & Census & m feat & Optdigits & CNAE-9 \\
            \hline
            Features      & 562  & 16    & 16   & 67    & 515 & 62 & 856   \\
            \hline
            Known classes & 3    & 19    & 5    & 12    & 5   & 5    & 4   \\
            Known data    & 3733 & 10229 & 3777 & 12000 & 802 & 1918 & 377 \\
            \hline
            Novel classes & 3    & 7     & 5    & 6     & 5   & 5    & 5   \\
            Novel data    & 3619 & 3770  & 3717 & 6000  & 798 & 1905 & 487 \\
            \hline
            Test data     & 1453 & 1704  & 1734 & 6000  & 202 & 905  & 113 \\
            \hline
        \end{tabular}
\end{table}

\textbf{Metrics.} We report the \textit{clustering accuracy} (ACC) on the unlabeled data.
It is defined as:
\begin{equation}
    ACC = \max_{perm \in P} \frac{1}{M} \sum_{i=1}^{M} \mathds{1} \left\{ y_i = perm(\hat{y}_i \right\}
\end{equation}
where $y_i$ and $\hat{y}_i$ are the ground truth labels and predicted labels for instance $x_i$ respectively.
Here, $M = \left| D^u \right|$.
$P$ is the set of all possible permutations between ground truth and predicted labels.
It can be easily computed using the Hungarian algorithm \cite{Kuhn55hungarian}.
The Normalized Mutual Information (NMI) and Adjusted Rand Index (ARI) are also reported in Appendix~\ref{sec:appendix_metrics}.

\textbf{Competitors.}
We report the performance of $k$-means and Spectral Clustering, along with their NCD-adapted versions introduced in Sections \ref{sec:ncdkmeans} and \ref{sec:ncdsc}.
We also include PBN and the pioneering NCD work for tabular data, TabularNCD \cite{tr2022method}.
Finally, we implement the same baseline used in \cite{tr2022method}.
It is a simple deep classifier that is first trained to distinguish the known classes.
Then, the penultimate layer is used to project the data of the novel classes before clustering it with $k$-means.
See the discussion of Section~\ref{sec:pbn} for more details.
We will call this approach the ``baseline'' for the remainder of this article.

\textbf{Implementation details.}
The neural-network based methods (i.e. PBN, TabularNCD and the baseline) are trained with the same architecture: an encoder of 2 hidden layers of decreasing size, and a final layer for the latent space whose dimension is optimized as an hyperparameter.
The dropout probability and learning rate are also hyperparameters to be optimized.
The classification networks are all a single linear layer followed by a softmax layer.
All methods are trained for 200 epochs and with a fixed batch size of 512.
For a fair comparison, the hyperparameters of these neural-network based methods are optimized following the process described in Section~\ref{sec:hp} (whereas the parameters of NCD SC are simply optimized following Section~\ref{sec:ncdsc}).
Thus, the labels of the novel classes are never used except for the computation of the evaluation metrics reported in the result tables.
The hyperparameters of the deep-based methods are tuned by a random search of the hyperparameter space and selecting the combination which obtained the best ARI on the hidden classes.
The search space and selected values can be found in Appendix~\hbox{\ref{sec:appendix_params}}.

\textbf{Objectives of the experiments.}
In the following section, an extensive evaluation of the clustering accuracy of the 7 competitors is performed.
This evaluation is carried out in two steps.
First, when the number of novel classes is known in advance, we seek to determine whether (1) the NCD-adapted versions of $k$-means and Spectral Clustering outperform their purely unsupervised versions when labeled data is available, and (2) which of the deep-based NCD methods performs best when the labels of the novel classes are not available for hyperparameter tuning.
And second, in the most realistic scenario where the number of classes is not given in advance, we compare the ability of different CVIs to accurately estimate this number and the impact on the NCD methods of using this estimation.

\subsection{Results analysis}

\subsubsection{Results when the number of novel classes is known in advance}
\label{sec:res1}
~\par
\textbf{Clustering.}
In Table~\ref{table:clust}, we first examine the performance of the unsupervised clustering methods when the number of novel classes $C^u$ is known in advance.
The aim is to determine which of the clustering algorithms performs best and should be compared with the NCD methods.
We observe that NCD $k$-means is never worse than $k$-means, and NCD SC is only once worse than SC.
This result confirms the efficacy of both NCD approaches and demonstrates that even simple clustering techniques can benefit from the known classes, although the improvements are sometimes only marginal.

The comparison between NCD $k$-means and NCD SC confirms the idea that no single clustering algorithm is universally better than the others in all scenarios, as noted by \cite{von2012clustering}.
However, NCD SC outperforms its competitors on 4 occasions and has a the highest average accuracy.
Therefore, this algorithm is selected for the next step of comparisons.

\begin{table}[htbp]
    \caption{Test ACC of the clustering algorithms averaged over 10 runs.}
    \label{table:clust}
    \centering
    \begin{tabular}{| l | c | c | c | c |}
        \hline
        \multirow{2}{*}{Dataset} & \multirow{2}{*}{$k$-means} & NCD       & \multirow{2}{*}{SC} & \multirow{2}{*}{NCD SC} \\
                                 &                            & $k$-means &                     &                         \\
        \hline
        Human      & 75.7$\pm$0.2 & 75.9$\pm$0.0 & 76.3$\pm$0.3 & \textbf{93.1$\pm$9.7} \\
        Letter     & 50.7$\pm$0.2 & 51.9$\pm$2.3 & 55.9$\pm$0.0 & \textbf{57.4$\pm$5.8} \\
        Pendigits  & 81.7$\pm$0.0 & 81.7$\pm$0.0 & \textbf{83.0$\pm$0.0} & 81.7$\pm$2.7 \\
        Census     & 49.9$\pm$4.0 & \textbf{50.4$\pm$1.1} & 48.5$\pm$0.3 & 48.0$\pm$1.8 \\
        m feat     & 89.1$\pm$0.3 & \textbf{89.7$\pm$0.4} & 89.6$\pm$0.3 & 89.2$\pm$2.3 \\
        Optdigits  & 79.1$\pm$4.5 & 94.2$\pm$0.0 & 89.7$\pm$0.0 & \textbf{95.4$\pm$5.3} \\
        CNAE-9     & 60.6$\pm$5.9 & 61.2$\pm$4.5 & 53.8$\pm$4.8 & \textbf{69.0$\pm$6.7} \\
        \hline
        Average    & 69.5 & 72.1 & 71.0 & \textbf{76.3} \\
        \hline
    \end{tabular}
\end{table}

\textbf{NCD.}
As shown in Table~\ref{table:ncd}, PBN outperforms both the baseline and TabularNCD by an average of 21.6\% and 12.9\%, respectively.
It is only outperformed by the baseline on the \textit{Letter Recognition} \cite{letterrecog} dataset.
This dataset consists of primitive numeric attributes describing the 26 capital letters in the English alphabet, which suggests a high feature correlation between the features used to distinguish the known and novel classes.
Since the baseline learns a latent space that is strongly discriminative for the known classes, this gives the baseline model a distinct advantage in this specific context.
On the other hand, we observe that it is at a disadvantage when the datasets do not share as many high-level features between the known and novel classes.

Table~\ref{table:ncd} also demonstrates the remarkable competitiveness of the NCD Spectral Clustering method, despite its low complexity.
On average, it trails behind PBN by only 1.0\% in ACC and manages to outperform PBN twice over 7 datasets.

\begin{table}[htbp]
    \caption{Test ACC of the NCD methods averaged over 10 runs.}
    \label{table:ncd}
    \centering
    \begin{tabular}{| l | c | c | c | c |}
        \hline
        Dataset & Baseline & NCD SC & TabularNCD & PBN \\
        \hline
        Human      & 71.6$\pm$1.7 & \textbf{93.1$\pm$9.7} & 72.2$\pm$2.6 & 76.7$\pm$1.8 \\
        Letter     & \textbf{64.9$\pm$2.6} & 57.4$\pm$5.8 & 62.1$\pm$3.0 & 62.4$\pm$2.0 \\
        Pendigits  & 53.4$\pm$6.6 & 81.7$\pm$2.3 & 57.0$\pm$6.0 & \textbf{82.8$\pm$0.6} \\
        Census     & 59.1$\pm$0.8 & 48.0$\pm$1.8 & 45.2$\pm$4.8 & \textbf{62.4$\pm$0.9} \\
        m feat     & 66.7$\pm$4.1 & 89.2$\pm$2.3 & 90.2$\pm$2.7 & \textbf{91.7$\pm$0.8} \\
        Optdigits  & 40.7$\pm$5.1 & \textbf{95.4$\pm$5.3} & 73.0$\pm$8.4 & 92.6$\pm$2.3 \\
        CNAE-9     & 40.2$\pm$3.2 & 69.0$\pm$6.7 & 51.3$\pm$5.2 & \textbf{72.6$\pm$4.6} \\
        \hline
        Average    & 55.7 & 76.3 & 64.4 & \textbf{77.3} \\
        \hline
    \end{tabular}
\end{table}

\begin{figure}[htbp]
    \hspace*{\fill}
    \begin{subfigure}{0.30\textwidth}
        \centering
        \includegraphics[width=1\textwidth]{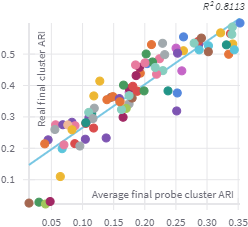}
        \caption{PBN}
        \label{fig:letter1}
    \end{subfigure}
    \hfill
    \begin{subfigure}{0.30\textwidth}
        \centering
        \includegraphics[width=1\textwidth]{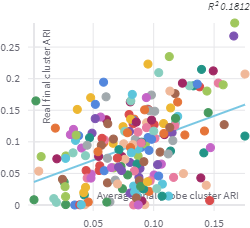}
        \caption{TabularNCD}
        \label{fig:letter2}
    \end{subfigure}
    \hspace*{\fill}
    \caption{Comparison between the ARI on the hidden and novel classes. Each point is a different hyperparameter combination.}
    \label{fig:letter_corr}
\end{figure}

To investigate the reasons behind the subpar performance of TabularNCD, we look at the correlation between the ARI on the hidden classes and the final ARI of the model on the novel classes.
A strong correlation would imply that if a combination of hyperparameters performed well on the hidden classes, it would also perform well on the novel classes.
To examine this, we plot the average ARI on the hidden classes against the ARI on the novel classes.
Figure~\ref{fig:letter_corr} is an example of such a plot.
It shows that, in the case of the \textit{Letter Recognition} dataset, PBN has a much stronger correlation than TabularNCD.
We attribute this difference to the large number of hyperparameters of TabularNCD (7, against 4 for PBN), which causes the method to overfit on the hidden classes, resulting in a lack of effective transfer of hyperparameters to the novel classes.

In conclusion, this section has shown that when the number $C^u$ of novel classes is known, NCD SC performs almost as well as PBN.
Therefore, in this specific scenario, NCD SC is a viable candidate for addressing the NCD problem due to its lower complexity and shorter training time.
Conversely, despite its strong learning capacity, TabularNCD is penalized by its high number of hyperparameters.
Finally, the performance of the methods as the number of novel classes increases is studied in Appendix~\hbox{\ref{sec:appendix_perf_wrt_share}}.
It shows that PBN is robust over a wide range of values, and that deep-based methods tend to fail when the number of known classes is too small.

\subsubsection{Results when the number of novel classes is estimated}
\label{sec:res2}
~\\
As expressed in Section~\ref{sec:est}, we leverage the representation learned by PBN during the $k$-fold CV to estimate the number of clusters.
The result is a method that never relies on any kind of knowledge from the novel classes.
This approach is also applicable to the baseline and the spectral embedding of SC, but not to TabularNCD as it requires a number of clusters to be defined during the training of its representation.
For a fair comparison, TabularNCD is trained here with a number of clusters that was estimated beforehand with a CVI.

\begin{figure}[htbp]
    \centering
    \includegraphics[width=0.65\textwidth]{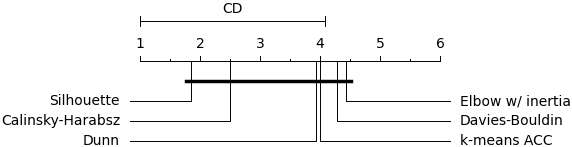}
    \caption{Comparison of the CVIs in the latent space of PBN using the Nemenyi test with a 95\% confidence interval.}
    \label{fig:cd_latent}
\end{figure}

To determine which CVI will perform the best in this application, we estimate the number of classes in the latent spaces learned by PBN when $C^u$ was known in advance.
Figure~\ref{fig:cd_latent} displays the average ranks of the CVIs in the latent spaces, and the details of the results can be found in Appendix~\ref{sec:appendix_k_est}.
The Nemenyi post-hoc test was used to compare the methods against all others.
However, given the relatively small number of datasets, the Critical Difference (CD) is large and the CVIs are not statistically different from each another according to the Nemenyi test.

Nevertheless, we find that the Silhouette coefficient performed the best in the latent space of PBN, closely followed by the Calinski-Harabasz index.
The elbow method is ranked last, which could be explained by the difficulty in defining an elbow
\footnote{There are no widely accepted approaches, as the concept of an ``elbow'' is subjective. 
In this study, we employed the \textit{kneedle} algorithm \cite{satopaa2011finding} through the \textit{kneed} Python Library \cite{arvai_kevin_2023_7873825}.}.

To summarize, in the following results, the Silhouette coefficient will be used for all estimations of $C^u$.
The full training procedure described in Section~\ref{sec:full_proce} will be used to train the baseline and PBN, with $C^u$ estimated in their latent spaces as detailed in Algorithm \ref{alg:cap}.
For TabularNCD, $C^u$ is estimated once in the original feature space (see Appendix~\ref{sec:appendix_k_est} Table~\ref{table:silhouette_results} for the values used).
And NCD SC is trained as it was described in Section~\ref{sec:ncdsc}, but with $C^u$ estimated in its spectral embedding.

\begin{table}[htbp]
    \caption{Test ACC averaged over 10 runs. With $C^u$ estimated with the Silhouette coefficient.}
    \label{table:ncd_k_est}
    \centering
    \begin{tabular}{| l | c | c | c | c |}
        \hline
        Dataset & Baseline & NCD SC & TabularNCD & PBN \\
        \hline
        Human      & 70.8$\pm$2.9 & 30.2$\pm$4.2 & \textbf{71.1$\pm$0.0} & \textbf{71.1$\pm$0.0} \\
        Letter     & \textbf{64.0$\pm$6.1} & 34.8$\pm$2.3 & 41.8$\pm$4.9 & 61.3$\pm$4.7 \\
        Pendigits  & 46.7$\pm$3.6 & 74.1$\pm$2.1 & 57.0$\pm$6.0 & \textbf{83.0$\pm$0.3} \\
        Census     & \textbf{56.6$\pm$3.6} & 29.0$\pm$3.5 & 35.7$\pm$1.6 & 49.8$\pm$0.1 \\
        m feat     & 59.5$\pm$7.7 & 73.2$\pm$3.9 & 41.1$\pm$0.2 & \textbf{90.6$\pm$2.2} \\
        Optdigits  & 42.1$\pm$4.6 & 79.5$\pm$4.0 & \textbf{96.9$\pm$1.3} & 90.5$\pm$4.8 \\
        CNAE-9     & 33.8$\pm$3.8 & 44.6$\pm$3.9 & 39.3$\pm$0.2 & \textbf{50.8$\pm$1.5} \\
        \hline
        Average    & 53.4 & 52.2 & 54.7 & \textbf{71.0} \\
        \hline
    \end{tabular}
\end{table}

As emphasised earlier, Table~\ref{table:ncd_k_est} reports the results of the different NCD algorithms in the most realistic scenario possible, where both the labels and the number of novel classes are not known in advance.
This is the setting where PBN exhibits the greatest improvement in performance compared to the other competitors, achieving an ACC that is 17.6\%, 18.8\% and 16.3\% higher than the baseline, NCD SC, and TabularNCD, respectively.

Remarkably, TabularNCD outperforms PBN on the Optdigits datasets where the number of clusters was overestimated by the Silhouette coefficient in the original feature space.
This suggests that TabularNCD probably only utilized the output neurons necessary for clustering, leaving the others unused, which was the method proposed in \cite{hsu2018learning} for estimating $C^u$.
This, however, is not true for the Letter dataset where $C^u$ was significantly overestimated, indicating that accurate estimations will likely result in improved performance.

Compared to the case where $C^u$ is known in advance, the ACC of the baseline falls from 55.7\% to 53.4\% and NCD SC falls from 74.3\% to 52.2\%.
This shows that they are both unable to find a latent space suitable for the estimation of $C^u$.

However, the ACC of PBN remains an impressive 71.0\%, demonstrating that this simple method, comprising of only two loss terms, is the most appropriate for tackling the NCD problem in a realistic scenario.
In contrast to the baseline and NCD SC methods, its reconstruction term enables it find a latent space where the unlabeled data are correctly represented.
And unlike TabularNCD, it has a low number of hyperparameters which decreases the probability of overfitting on the hidden classes during the $k$-fold CV procedure.

\section{Conclusion}

In this article, we have shown that in the NCD setting, unsupervised clustering algorithms can benefit from knowledge of the known classes and reliably improve their performance by implementing simple modifications.
We have also introduced a novel NCD algorithm called PBN, which is characterized by its simplicity and low number of hyperparameters, which proved to be a decisive advantage under realistic conditions.
In addition, we have proposed an adaptation of the $k$-fold cross-validation process to tune the hyperparameters of NCD methods without depending on the labels of the novel classes.
Finally, we have demonstrated that the number of novel classes can be accurately estimated within the latent space of PBN.
These two previous contributions have shown that the NCD problem can be solved in realistic situations where no prior knowledge of the novel classes is available during training.


\section*{Declarations}

\subsection*{Funding}
Colin Troisemaine, Alexandre Reiffers-Masson, St\'ephane Gosselin, Vincent Lemaire and Sandrine Vaton received funding from Orange SA.

\subsection*{Competing Interests}
Colin Troisemaine, St\'ephane Gosselin and Vincent Lemaire received research support from Orange SA.
Alexandre Reiffers-Masson and Sandrine Vaton received research support from IMT Atlantique.

\subsection*{Ethics approval}
Not applicable.

\subsection*{Consent to participate}
All authors have read and approved the final manuscript.

\subsection*{Consent for publication}
Not applicable.

\subsection*{Availability of data and materials}
All data used is this study are available publicly online. The datasets were extracted directly in the repositories available with the links in the corresponding section.

\subsection*{Code availability}
The code for experiments is available at the following url: \url{https://github.com/Orange-OpenSource/PracticalNCD}.

\subsection*{Authors' contributions}
Colin Troisemaine, Alexandre Reiffers-Masson, St\'ephane Gosselin, Vincent Lemaire and Sandrine Vaton contributed to the manuscript equally.


\bibliography{bibliography}

\clearpage

\begin{appendices}

\section{Pseudocode of the methodologies}
\label{sec:appendix_pseudocode}

In this section, we provide the pseudocode of the NCD $k$-means and NCD Spectral Clustering methods proposed in Sections \hbox{\ref{sec:ncdkmeans}} and \hbox{\ref{sec:ncdsc}}, respectively.
With this addition, we hope to improve the clarity of our research and make it easier to understand the available open-source code.

\newcommand\Algphase[1]{%
\vspace*{-.7\baselineskip}\Statex\hspace*{\dimexpr-\algorithmicindent-2pt\relax}\rule{\textwidth}{0.4pt}%
\Statex\hspace*{-\algorithmicindent}\textit{#1}%
\vspace*{-.7\baselineskip}\Statex\hspace*{\dimexpr-\algorithmicindent-2pt\relax}\rule{\textwidth}{0.4pt}%
}

\begin{algorithm}[htbp]
    \caption{NCD $k$-means}
    \label{alg:ncdkmeans}
    \begin{algorithmic}[1]
        \Require Labeled set of known classes $D^l = \{ X^l, Y^l \}$, unlabeled set of novel classes $D^u = \{ X^u \}$, number of clusters $k$.
        \State \textbf{Initialize:} $L^l$ and $L^u$ empty lists of centroids for the known and novel classes
        \Algphase{Phase 1 - Initialization of the centroids}
        \State \textit{// Initialize the centroids of known classes using the ground truth labels:}
        \For{each class $c$ in $D^l$}
            \State Let $X^l_{c} = \{ x_i^l \mid x_i^l \in X^l, y_i^l = c \}$ be the subset of points belonging to class $c$
            \State Calculate $\mu_{c}^l = \frac{1}{| X^l_{c} |}\sum_{x_i^l \in X^l_{c}} x_i^l$ the average point of class $c$
            \State Add $\mu_{c}^l$ to the list $L^l$ of centroids of the known classes
        \EndFor
        \State \textit{// Initialize the centroids of the novel classes in $L^u$ following $k$-means++:}
        \While{$| L^u | < k$}
            \State For each point $x_i^u$ in $D^u$, calculate the distance $d(x_i^u)$ between $x^u_i$ and its nearest centroid in $L^l \cup L^u$
            \State Choose the next centroid from the points in $D^u$ with probability proportional to $d(x_i^u)^2$
            \State Add the chosen centroid to $L^u$
        \EndWhile
        \Algphase{Phase 2 - Follow the usual $k$-means algorithm}
        \State Make the centroids of $L^u$ converge, using only the data of $D^u$
        \State Partition the points in $D^u$ using only the centroids of the novel classes in $L^u$
    \end{algorithmic}
\end{algorithm}

\begin{algorithm}[htbp]
    \caption{NCD Spectral Clustering}
    \label{alg:ncdsc}
    \begin{algorithmic}[1]
        \Require Labeled set of known classes $D^l = \{ X^l, Y^l \}$, unlabeled set of novel classes $D^u = \{ X^u \}$, number of clusters $k$, number of hyperparameter optimization runs $n_{opt}$.
        \Algphase{Phase 1 - Determination of the optimal hyperparameters}
        \State Initialize $ARI_{opt} = 0$ the best ARI obtained and $\{ u_{opt} = 0, \sigma_{opt} = 0 \}$ the optimal hyperparameters
        \For{$n_{opt}$ random combinations of hyperparameters $\{ u_{rand}, \sigma_{rand} \}$}
            \State Compute the adjacency matrix $A$ of the points in $D^l \cup D^u$, where $A_{i,j} = \text{exp}(- \lVert x_i - x_j \rVert^2_2 / (2\sigma_{rand}^2))$
            \State Define the spectral embedding $U$ as the first $u_{rand}$ eigenvectors of the symmetric normalized Laplacian of $A$
            \State Partition all the points in $U$ using $k$-means into $k + C^l$ classes
            \State Calculate $ARI_{rand}$, the clustering ARI of this $k$-mean only on the points of $D^l$, using the ground truth labels $Y^l$
            \If{$ARI_{rand} > ARI_{opt}$}
                \State $ARI_{opt} \gets ARI_{rand}$, $u_{opt} \gets u_{rand}$ and $\sigma_{opt} \gets \sigma_{rand}$
            \EndIf
        \EndFor
        \Algphase{Phase 2 - Clustering of the data of the novel classes}
        \State Get $U_{opt}$ the spectral embedding of $D^l \cup D^u$ with the parameters $\{ u_{opt}, \sigma_{opt} \}$
        \State Partition the points of $D^u$ embedded in $U_{opt}$ into $k$ clusters with $k$-means
    \end{algorithmic}
\end{algorithm}

\newpage

\section{Additional result metrics}
\label{sec:appendix_metrics}

In addition to the ACC results discussed in Section~\ref{sec:res2}, we present the NMI (Table~\ref{table:final_table_nmi}) and ARI (Table~\ref{table:final_table_ari}) for the NCD methods when the number of novel classes $C^u$ is unknown and has to be estimated.
Our results are consistent with those shown in Table~\ref{table:ncd_k_est}:
the PBN method largely outperforms its competitors in this realistic scenario.
In particular, PBN achieves an average NMI higher than the baseline, NCD SC and TabularNCD by 23.5\%, 8.6\% and 13.5\% respectively.
Similarly, the ARI is 24.5\%, 12.4\% and 17.5\% higher on average.

\begin{table}[htbp]
    \caption{Test NMI averaged over 10 runs. With $C^u$ estimated with the Silhouette coefficient.}
    \label{table:final_table_nmi}
    \centering
    \begin{tabular}{| l | c | c | c | c |}
        \hline
        Dataset & Baseline & NCD SC & TabularNCD & PBN \\
        \hline
        Human     & 56.1$\pm$8.9 & 45.8$\pm$3.0 & \textbf{75.2$\pm$0.0} & \textbf{75.2$\pm$0.0}  \\
        Letter    & 54.4$\pm$3.3 & 52.6$\pm$1.0 & 37.2$\pm$3.2 & \textbf{59.2$\pm$2.4 } \\
        Pendigits & 41.0$\pm$6.7 & \textbf{79.5$\pm$1.4} & 48.0$\pm$6.5 & 73.4$\pm$2.5  \\
        Census    & 59.7$\pm$0.6 & 33.5$\pm$1.0 & 42.6$\pm$1.5 & \textbf{60.6$\pm$0.5}  \\
        m feat    & 50.0$\pm$3.9 & 71.9$\pm$2.9 & 44.8$\pm$2.1 & \textbf{79.1$\pm$3.3}  \\
        Optdigits & 28.5$\pm$6.0 & 77.3$\pm$2.0 & \textbf{93.1$\pm$2.0} & 84.9$\pm$2.0  \\
        CNAE-9    & 23.0$\pm$1.5 & \textbf{56.5$\pm$2.0} & 42.0$\pm$1.3 & 45.2$\pm$13.4  \\
        \hline
        Average   & 44.7 & 59.6 & 54.7 & \textbf{68.2} \\
        \hline
    \end{tabular}
\end{table}

\begin{table}[htbp]
    \caption{Test ARI averaged over 10 runs. With $C^u$ estimated with the Silhouette coefficient.}
    \label{table:final_table_ari}
    \centering
    \begin{tabular}{| l | c | c | c | c |}
        \hline
        Dataset & Baseline & NCD SC & TabularNCD & PBN \\
        \hline
        Human     & 48.7$\pm$4.0 & 20.3$\pm$3.0 & \textbf{61.4$\pm$0.0} & \textbf{61.4$\pm$0.0} \\
        Letter    & 44.3$\pm$4.1 & 29.5$\pm$2.2 & 23.4$\pm$5.6 & \textbf{48.9$\pm$3.1} \\
        Pendigits & 29.8$\pm$5.7 & \textbf{73.7$\pm$1.7} & 37.6$\pm$6.5 & 65.3$\pm$3.4 \\
        Census    & \textbf{42.6$\pm$3.7} & 23.0$\pm$3.8 & 26.7$\pm$0.8 & 35.5$\pm$0.2 \\
        m feat    & 40.9$\pm$5.3 & 64.1$\pm$4.5 & 21.5$\pm$0.2 & \textbf{79.1$\pm$4.2} \\
        Optdigits & 16.9$\pm$6.5 & 75.6$\pm$3.6 & \textbf{94.1$\pm$2.7} & 84.4$\pm$4.4 \\
        CNAE-9    & 11.6$\pm$1.9 & \textbf{33.2$\pm$4.0} & 18.8$\pm$0.4 & 31.5$\pm$1.7 \\
        \hline
        Average   & 33.5 & 45.6 & 40.5 & \textbf{58.0} \\
        \hline
    \end{tabular}
\end{table}

\section{Impact of the ratio of novel classes}
\label{sec:appendix_perf_wrt_share}

\begin{figure}[htbp]
    \centering
    \includegraphics[width=0.85\textwidth]{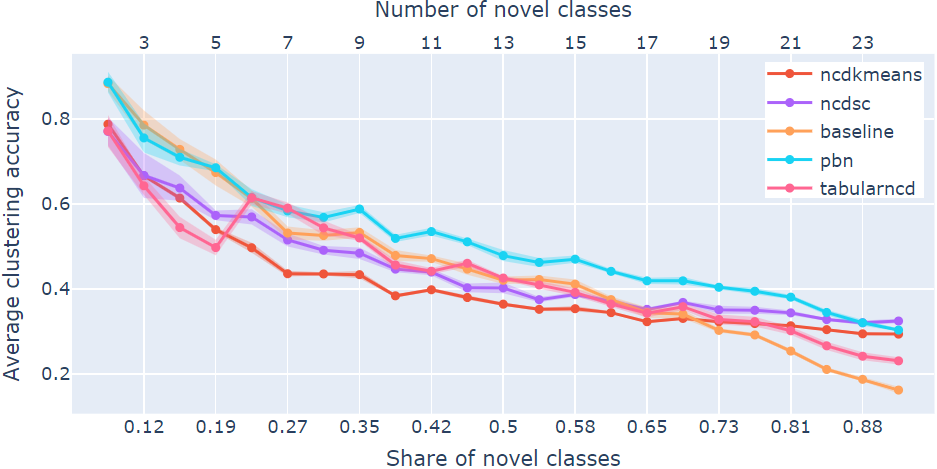}
    \caption{Comparison of the performance of the 5 methods discussed in this paper when the number of novel classes increases.}
    \label{fig:perf_wrt_share}
\end{figure}

In the literature on Novel Class Discovery, the ratio of the number of novel classes to the number of known classes is usually set arbitrarily (e.g. $0.5$ if there are few classes, and $0.2$ if there are many).
In reality, this ratio is not known in advance, so NCD methods should ideally be robust over a wide range of ratios.
To compare the robustness of the 5 methods discussed in this paper, we perform experiments on the \textit{Letter Recognition} \hbox{\cite{letterrecog}} dataset when the number of novel classes increases.
For each number of novel classes that was evaluated, we defined 5 random combinations of known/novel classes.
For example, in Figure~\hbox{\ref{fig:perf_wrt_share}}, the second point from the left corresponds to the average performance of the methods over 5 random combinations of 3 novel and 23 known classes.

From the results displayed in Figure~\hbox{\ref{fig:perf_wrt_share}}, we find that the performance of the methods is almost always uniformly decreasing as the ratio of novel classes increases.
It can also be observed that the accuracy of the deep-based methods (PBN, TabularNCD and the baseline) start decreasing faster than NCD $k$-means and NCD SC as the number of novel classes becomes too large.
As the number of known classes $C^l$ decreases, the quality of their latent representation is reduced and their clustering is negatively affected.
Of course, this is especially true for the baseline, which trains its latent space solely on the known classes.

The performance of TabularNCD peaks around 7 novel classes, which is the number of novel classes for which it was specifically optimized.
On the other hand, PBN is more stable and outperforms the competitors over a wide range of values.
Its small number of hyperparameters, coupled with its reconstruction loss, makes it less prone to overfitting, and it still extracts relevant information into its latent space even when there are few known classes.

\section{Hyperparameters details}
\label{sec:appendix_params}

The Table~\ref{table:fair_basleine_hyperparams_k_estim} shows the hyperparameters found by the full procedure described in Section~\ref{sec:full_proce}.
The type of distributions and value spaces explored for each hyperparameter are given in the ``Range'' and ``Distribution'' columns.
The variable $d$ in the range of the \textit{latent dim} refers to the dimension of the points in the datasets.

\begin{table}[htbp]
    \caption{Hyperparameters of the methods found when $C^u$ is estimated.}
    \label{table:fair_basleine_hyperparams_k_estim}
    \centering
    \footnotesize
    \setlength{\tabcolsep}{1pt}
        \begin{tabular}{c | l | c | c | c | c | c | c | c | c | c }
            \hline
            \multicolumn{2}{c|}{Parameter} & Range & Distribution & Human & Letter & Pendigits & Census & m features & Optdigits & CNAE-9 \\
            \hline
            \multirow{3}{*}{\rotatebox[origin=c]{90}{\parbox[c]{8mm}{\centering Base-line}}}
                & latent dim    & [5, $d$]      & int uniform & 560      & 9        & 9        & 30       & 34       & 42       & 569      \\
                & lr            & [0.0001, 0.1] & log uniform & 0.000154 & 0.001333 & 0.005517 & 0.002021 & 0.000118 & 0.003330 & 0.000167 \\
                & dropout       & [0.0, 0.6]    & uniform     & 0.168044 & 0.140095 & 0.052505 & 0.467836 & 0.564855 & 0.075180 & 0.257765 \\
            \hline
            \multirow{2}{*}{\rotatebox[origin=c]{90}{\parbox[c]{6mm}{\centering NCD SC}}}
                & $s_{min}$     & ]0.0, 1.0[    & uniform     & 0.29396 & 0.98137 & 0.86147 & 0.63534 & 0.70005 & 0.20412 & 0.83132 \\
                & $u$           & [1, 200]      & int uniform & 144     & 14      & 18      & 53      & 16      & 26      & 14      \\
            \hline
            \multirow{7}{*}{\rotatebox[origin=c]{90}{TabularNCD}}
                & $k$ neighbors & [2, 100]      & int uniform & 70      & 29      & 73      & 45      & 15      & 48      & 7       \\
                & latent dim    & [5, $d$]      & int uniform & 372     & 14      & 13      & 56      & 98      & 62      & 791     \\
                & lr            & [0.0001, 0.1] & log uniform & 0.00961 & 0.00036 & 0.00464 & 0.00124 & 0.00840 & 0.00755 & 0.00129 \\
                & dropout       & [0.0, 0.6]    & uniform     & 0.15436 & 0.06672 & 0.27823 & 0.27489 & 0.01384 & 0.09158 & 0.08220 \\
                & top $k$       & [0.0, 1.0]    & uniform     & 0.61097 & 0.16517 & 0.46739 & 0.93956 & 0.59044 & 0.99586 & 0.94210 \\
                & $w_1$         & [0.0, 1.0]    & uniform     & 0.04745 & 0.41367 & 0.13992 & 0.31247 & 0.30193 & 0.78532 & 0.19631 \\
                & $w_2$         & [0.0, 1.0]    & uniform     & 0.70265 & 0.97095 & 0.60544 & 0.88265 & 0.83427 & 0.93356 & 0.64128 \\
            \hline
            \multirow{4}{*}{\rotatebox[origin=c]{90}{PBN}}
                & latent dim    & [5, $d$]      & int uniform & 504     & 22      & 12      & 13      & 172     & 53      & 579     \\
                & lr            & [0.0001, 0.1] & log uniform & 0.00010 & 0.00058 & 0.00107 & 0.00211 & 0.00467 & 0.00036 & 0.00440 \\
                & dropout       & [0.0, 0.6]    & uniform     & 0.50984 & 0.02745 & 0.01126 & 0.16777 & 0.23934 & 0.06606 & 0.08131 \\
                & $w$           & [0.0, 1.0]    & uniform     & 0.46714 & 0.72241 & 0.10671 & 0.96086 & 0.67034 & 0.18917 & 0.69815 \\
            \hline
        \end{tabular}
\end{table}

\section{PBN coupled with Spectral Clustering}
\label{sec:appendix_pbn_sc}

In Section~\hbox{\ref{sec:ncdsc}}, we proposed the Projection Based NCD (PBN) method where a latent representation is first trained with a reconstruction loss on all the points, and a classification loss on the labeled data.
After training of the representation, the simple $k$-means algorithm is used to cluster the unlabeled data of the novel classes.
It was chosen mainly for its fast execution time, as the idea behind PBN is to obtain a representation where the novel classes are already well separated, so a sophisticated clustering algorithm should not be needed.
But naturally, any other unsupervised clustering method can be employed and might improve the performance.
Therefore, in this section, we investigate the performance of PBN when $k$-means is replaced with Spectral Clustering (SC).

In these experiments, the number of novel classes $C^u$ is considered to be known in advance.
In Table~\hbox{\ref{table:pbn_sc}}, three variations of the PBN method are compared:
\begin{itemize}
    \item \textbf{KM PBN:} The model used in the article, where $k$-means is used in the latent space.
    \item \textbf{SC PBN default:} $k$-means is directly replaced by SC with default hyperparameters.
    The similarity $s_{min}$ is set to 0.6, and the number of components $u$ of the spectral embedding is set to the number of novel classes $C^u$, as is usually done in the literature on Spectral Clustering.
    \item \textbf{SC PBN full opt:} Here, SC is also used for clustering, and its hyperparameters $s_{min}$ and $u$ are optimized together with PBN's hyperparameters. So for each datasets, 6 hyperparameters are optimized (4 from PBN and 2 from SC).
\end{itemize}

\begin{table}[htbp]
    \centering
    \caption{Test accuracy of the 3 variations of PBN averaged over 10 runs.}
    \label{table:pbn_sc}
        \begin{tabular}{| l | c | c | c |}
            \hline
            \multirow{2}{*}{Dataset} & \multirow{2}{*}{KM PBN} & SC PBN & SC PBN \\
            & & default & full opt\\
            \hline
            Human     & 76.7$\pm$1.8 & \textbf{80.2$\pm$0.7} & 76.6$\pm$6.1 \\
            Letter    & 62.4$\pm$2.0 & \textbf{64.4$\pm$2.8} & 50.9$\pm$2.7 \\
            Pendigits & \textbf{82.8$\pm$0.6} & 82.0$\pm$2.2 & 79.3$\pm$4.0 \\
            Census    & \textbf{62.4$\pm$0.9} & 61.1$\pm$2.0 & 53.0$\pm$2.4 \\
            m feat    & 91.7$\pm$0.8 & \textbf{92.6$\pm$1.6} & 89.7$\pm$1.5 \\
            Optdigits & 92.6$\pm$2.3 & \textbf{94.8$\pm$2.1} & 94.6$\pm$3.6 \\
            CNAE-9    & 72.6$\pm$4.6 & \textbf{73.5$\pm$6.1} & 68.0$\pm$4.2 \\
            \hline
            Average   & 77.3 & \textbf{78.4} & 73.2 \\
            \hline
        \end{tabular}
\end{table}

In Table~\hbox{\ref{table:pbn_sc}}, it can be seen that using SC with its default hyperparameters results in a clustering accuracy that is on average 1.1\% higher than PBN coupled with $k$-means.
But when the hyperparameters of SC are also optimized, the average accuracy drops by 4.1\% instead.
This shows that, as expressed throughout this article, having too many hyperparameters in the context of the NCD problem makes it difficult to effectively transfer the results on the known classes to the novel classes, and the hyperparameters tend to overfit the known classes.

These results suggest that the best approach is to apply SC with its default hyperparameters to cluster the projection of the novel data.
However, it must be noted that the number of novel classes is used to define the number of components of the spectral embedding.
Thus, in the case where $C^u$ is not considered to be known in advance and has to be estimated, the performance may be degraded.

\section{Cluster Validity Indices results details}
\label{sec:appendix_k_est}

An estimate of the number of clusters in the 7 datasets considered in this paper can be found in Table~\ref{table:silhouette_results}.
Among the 6 CVIs reported here, the Silhouette coefficient performed the best.
Furthermore, compared to the original feature space, its average estimation error significantly decreased in the latent space, validating our approach.
For some datasets, the Davies-Bouldin index continued to decrease and the Dunn index continued to increase as the number of clusters increased, resulting in very large overestimations.
Note that the estimates of the number of novel classes in Table \ref{table:silhouette_results} are not needed in the experiments of Section \ref{sec:res2}, since Algorithm \ref{alg:cap} directly incorporates such estimates in the training procedure.
This table has only helped us to identify the most appropriate CVI for our problem.
The only exception is the TabularNCD method, which requires an a priori estimation of the number of novel classes in the original feature space.

\begin{table}[htbp]
    \caption{An estimation of the number of novel classes with some CVIs in the latent space of PBN.}
    \label{table:silhouette_results}
    \centering
        \begin{tabular}{| l | c | c | c | c | c | c | c |}
            \hline
            Dataset & Human & Letter & Pendigits & Census & m feat & Optdigits & CNAE-9 \\
            \hline
            Ground-truth & 3 & 7 & 5 & 6 & 5 & 5 & 5 \\
            \hline
            \multicolumn{8}{c}{PBN latent space} \\
            \hline
            Silhouette & 2 & 8 & 5 & 3 & 5 & 5 & 5 \\
            \hline
            CH         & 2 & 3 & 5 & 4 & 2 & 2 & 5 \\
            \hline
            Dunn       & 2 & 3 & 98 & 3 & 2 & 95 & 2 \\
            \hline
            KM ACC     & 1 & 2 & 3 & 3 & 5 & 6 & 1 \\
            \hline
            Davies-B.  & 2 & 63 & 6 & 3 & 99 & 4 & 96 \\
            \hline
            Elbow      & 9 & 14 & 10 & 7 & 16 & 13 & 9 \\
            \hline
            \multicolumn{8}{c}{Original feature space} \\
            \hline
            Silhouette & 2 & 45 & 5 & 3 & 2 & 9 & 2 \\
            \hline
        \end{tabular}
\end{table}

\section{NCD k-means centroids convergence}
\label{sec:appendix_centroids}

In this appendix, we aim to determine how to achieve the best performance with NCD $k$-means.
Specifically, after the centroid initialization described in Section~\ref{sec:ncdkmeans}, we investigate:
(1) whether it is more effective to update the centroids of both known and novel classes, or only the centroids of novel classes;
(2) whether the centroids need to be updated using data from both known and novel classes, or only using data from novel classes.
The results are presented in Table~\ref{table:clust_results_different_convergence} and show that for 5 out of 7 datasets, the best results are obtained when only the centroids of the novel classes are updated on the unlabeled data.
Updating the centroids of the known classes always leads to worse performance, as the class labels are not used in this process.
Thus, the centroids of the known classes run the risk of capturing data from the novel classes (and vice versa).

\begin{table}[htbp]
    \caption{ACC of NCD $k$-means averaged over 10 runs.} 
    \label{table:clust_results_different_convergence}
    \centering
        \begin{tabular}{| l | l | c | c | c |}
            \hline
            \multirow{2}{*}{Dataset} & Converging the & On unlabeled & On labeled and \\
            & centroids... & data only & unlabeled data \\
            \hline
            \multirow{2}{*}{Human} & novel only      & 75.9$\pm$0.0 & \textbf{77.4$\pm$0.0} \\
                                   & known and novel & -            & 75.1$\pm$0.5          \\
            \hline
            \multirow{2}{*}{Letter} & novel only      & \textbf{51.9$\pm$2.3} & 39.5$\pm$1.9 \\
                                    & known and novel & -                     & 42.3$\pm$2.6 \\
            \hline
            \multirow{2}{*}{Pendigits} & novel only      & \textbf{81.7$\pm$0.0} & 72.7$\pm$0.9 \\
                                       & known and novel & -                     & 75.3$\pm$4.0 \\
            \hline
            \multirow{2}{*}{Census} & novel only      & \textbf{50.4$\pm$1.1} & 50.4$\pm$4.8 \\
                                    & known and novel & -                     & 44.6$\pm$8.3 \\
            \hline
            \multirow{2}{*}{m feat} & novel only      & \textbf{89.7$\pm$0.4} & 69.1$\pm$0.2 \\
                                    & known and novel & -                     & 84.1$\pm$7.0 \\
            \hline
            \multirow{2}{*}{Optdigits} & novel only      & \textbf{94.2$\pm$0.0} & 70.8$\pm$7.8  \\
                                       & known and novel & -                     & 74.0$\pm$14.7 \\
        \hline
            \multirow{2}{*}{CNAE-9} & novel only      & 61.2$\pm$4.5 & 48.3$\pm$8.5          \\
                                    & known and novel & -            & \textbf{68.1$\pm$7.5} \\
            \hline
        \end{tabular}
\end{table}

\end{appendices}

\end{document}